# ArNLI: Arabic Natural Language Inference for Entailment and Contradiction Detection

Khloud Al Jallad, Nada Ghneim

k-aljallad@aiu.edu.sy, n-ghneim@aiu.edu.sy

Faculty of Information Technology Engineering, Arab International University, Daraa, Syria

## ABSTRACT

Natural Language Inference (NLI) is a hot topic research in natural language processing, contradiction detection between sentences is a special case of NLI. This is considered a difficult NLP task which has a big influence when added as a component in many NLP applications, such as Question Answering Systems, text Summarization. Arabic Language is one of the most challenging low-resources languages in detecting contradictions due to its rich lexical, semantics ambiguity. We have created a data set of more than 12k sentences and named ArNLI, that will be publicly available. Moreover, we have applied a new model inspired by Stanford contradiction detection proposed solutions on English language. We proposed an approach to detect contradictions between pairs of sentences in Arabic language using contradiction vector combined with language model vector as an input to machine learning model. We analyzed results of different traditional machine learning classifiers and compared their results on our created data set (ArNLI) and on an automatic translation of both PHEME, SICK English data sets. Best results achieved using Random Forest classifier with an accuracy of 99%, 60%, 75% on PHEME, SICK and ArNLI respectively.

## KEYWORDS

Textual Entailment, Arabic NLP, Contradiction Detection, Contradiction Arabic Dataset, Textual Inference

## 1. INTRODUCTION

Natural language inference (NLI) is the task of determining whether a given hypothesis can be inferred from a given premise. This task, formerly known as recognizing textual entailment (RTE) has long been a popular task among researchers [1]. As an improvement over the simple binary Entailment vs Non-entailment scenario, three-way RTE has appeared and commonly used (Entailment, Contradiction, Neutral (Unknown)). The Entailment relation between two text fragments holds if the claim present in fragment B can be concluded from fragment A. The Contradiction relation applies when the claim in A and the claim in B cannot be true together. The Neutral relation applies if A and B neither entail nor contradict each other.

The main impact is that RTE can transfer problem from text data set language processing to algebra sets and logical implications, for that reason RTE has a big influence when added as a component in many NLP applications, as it can simplify problems.

Textual Inference is a key capability for improving performance in a wide range of NLP tasks [2], such as Question Answering Systems [3], Text Summarization[1][2], next-generation Information Retrieval [3], Machine Reading [4] [5], Machine Translation [6], Natural Language Understanding (NLU) [7], Anaphora Resolution [8] and Argumentation Mining [9].

Since 2005, several challenges have been coordinated with the aim of providing concrete data sets that the research community could use to test and compare their different approaches to recognize entailments.

However, RTE from Arabic text remains very little explored. Arabic Language is one of the most challenging low-resources languages in detecting contradictions due to its lexical richness and semantics ambiguity. Moreover, to the best of our knowledge there is no available benchmark for contradiction detection task in Arabic language.

In this paper, we introduce a new high quality data set for the NLI task for Arabic language. This data set, named ArNLI, includes more than 6000 pairs of sentences annotated in 3-way relation classes (entailment, contradiction, and neutral), where:

- **Contradiction** indicates contradict between two texts, involving all types of De Marneffe et al. discussed in [10] (Antonym, Negation, Numeric, Factive, Structure, Lexical, WK)
- **Entailment** indicates that two texts entail the same meaning.
- **Neutral** indicates that there is no relation between two texts.

Using different language modelling approaches (including word embeddings), and features of different language levels (lexical, semantic.), we evaluate different traditional classification models (Support Vector Machine (SVM), Stochastic Gradient Descent (SGD), Decision Tree (DT), ADA Boost, K-Nearest Neighbours (KNN), and Random Forest (RF)), and compare the results with translation of famous English benchmarks because of lack of benchmarks in Arabic.

The rest of the paper is organized as follows: Section 2 will cover the related literature. Section 3 will present our methodology in details, and Section 4 will describe our created Arabic RTE data set. Section 5 will then discuss the experiments results. Finally, in Section 6, we conclude with future research directions

## 2. RELATED WORKS

In the recent past, Natural language Inference (NLI) (formerly known as RTE) has gained significant attention, particularly given its promise for downstream NLP tasks [4]. The majority of researches done in NLI focus on two-way RTE (the simple binary Entailment vs Non-entailment scenario), whereas three-way RTE (Entailment, Contradiction, Neutral (Unknown)) that focus on contradiction has few number of researches. Recent statistics[3] shows that researches in RTE focus on big data sets using deep learning models with transformers such as BERTNLI, RoBERTa, XLNET, DeBERTa. Thus, most progress in NLI has been limited to English due to lack of reliable data sets for most of the world's languages. In other languages, different research works have attempted to create data sets for NLI such as for Japanese [11], Chinese [12], Portuguese [13], Italian [14], German [15], Brazilian [16] , Persian [17],and Turkish [18].

---

[1] NIST, "PASCAL Recognizing Textual Entailment Challenge (RTE-5) at TAC 2009,": https://tac.nist.gov/2009/RTE/

[2] NIST, "6th textual entailment challenge @ tac 2010 knowledge base population validation pilot task guidelines," TAC Workshop, 2010
[3] https://paperswithcode.com/sota/natural-language-inference-on-rte

As for Arabic language, although an Arabic data set for RTE[4] exists, but it convers two-way RTE, and has only 600 pairs, which is considered not enough for any deep learning methodology. In the rest of related works section, we will emphasis on three-way RTE that focus on contradiction has few number of researches, as it is our research interest in this paper.

## 2.1. Related Works on English Language

Harabagiu et al. [19], presented the first empirical results for contradiction detection (CD) as a task of entailment recognition, but they focused on specific kinds of contradiction and described a framework for detecting contradictions between sentences. The work has three basic types of linguistic information: (a) negation; (b) relational and modality features, and (c) semantic information. They created two corpora for evaluating their system. One was constructed via negating each entailment in the RTE2 data[5], generating a balanced data set (LCC1 negation data set). To keep away from overtraining, negative markers were also added to every non-entailment, making sure that they are not contradictions. The other corpus was created by paraphrasing the hypothesis sentences from LCC-negation to remove negations (LCC-paraphrase). They achieved accuracies of 75.63% on LCC-negation and 62.55% on LCC-paraphrase.

Rafferty and Manning in [10] proposed an appropriate definition of contradiction for NLP tasks and developed a corpus from which they constructed a typology of contradictions. They found two primary categories of contradiction: (1) those occurring via antonym, negation, and date/number mismatch, which are relatively simple to detect, and (2) contradictions arising from the use of factive or modal words, structural and subtle lexical contrasts, as well as world knowledge (WK). They considered contradictions in the first category 'easy' and can be obtained using existing resources and techniques (e.g., WordNet[6], VerbOcean). However, contradictions in the second category were considered more difficult to detect automatically because they require precise models of sentence meaning. Moreover ,they proposed a system based on the architecture of the Stanford RTE system [20] , however, they introduced a stage for event co-reference decision. The features used were: Polarity features, Number, date and time expression features, Antonym features, Structural features, Factivity corpora, one is based on RTE data set and the other is based on 'real life' data. As the RTE data sets are balanced between entailments and non-entailments, RTE3-test data was annotated by NIST as part of the RTE3 Pilot task[7] in which systems classify pairs as entailed, contradictory, or neither. As for real life corpus[8], they collected 131 contradictory pairs: 19 from newswire, mainly looking at related articles in Google News, 51 from Wikipedia, 10 from the Lexis Nexis database, and 51 from the data

---

prepared by LDC for the distillation task of the DARPA GALE program. Despite the randomness of the collection, they argued that this corpus may be best reflecting naturally occurring contradictions.

Ritter et al. [21], proposed Contradiction Detection using functions (e.g., BornIn (Person) = Place), and a domain-independent algorithm that automatically detects sentences denoting functions. Their work was based on de Marneffe et al.'s work with a number of modifications. They suggested that global world knowledge is important for constructing a domain-independent system. Moreover, they automatically created a large corpus of obvious contradictions found in arbitrary Web text. As for system evaluation, they used the 1,000 most frequent relations extracted by TextRunner system [22], 75% were indeed functional. They labelled by hand each of these 8,844 pairs as contradictory or not.

Li et al. in [23] used CNN-based (Convolutional Neural Network) model to learn the global and local semantic relation from sentences. They used contradiction-specific word embedding (CWE). CWE is learned from a training corpus that is automatically generated from the paraphrase database, and is used as features to implement contradiction detection in SemEval 2014 benchmark data set[9]. Shallow features extracted were: Number of negation words, Difference of word order, Unaligned words. Experimental results show optimization on traditional context-based word embedding in contradiction detection as it improved the accuracy from 75.97% to 82.08% in the contradiction class.

Sulea in [24] proposed to apply 3-way RTE in social media. The author worked on 5000 pairs collected from Twitter to distinguish between tweets that entail or contradict each other or that claim unrelated things. They used neural networks and compare their results on word embeddings with the results obtained previously using classical "feature engineering" methods.

Lingam et al. [25] proposed an approach for detecting three different types of contradiction: negation, antonyms and numeric mismatch using neural networks and deep learning. They used Long Short-Term Memory (LSTM) and Global Vectors for Word Representation (GloVe)[10] There are three feature combinations: manual features (Jaccard Coefficient, IsNegation Flag, IsAntonym Flag, Overlap Coefficient), LSTM based features and combination of manual and LSTM features. They did experiments on three publicly available data sets: Stanford data set, SemEval data set[11] and PHEME data set[12] [26]. In addition, they constructed a data set and made it publicly available. They achieved 96.85% accuracy for the contradiction class on the PHEME data set.

Moreover, in the last few years many research papers have applied NLI for special domains or to optimize solutions for other complex NLP tasks. For example, Microsoft created a corpus named Microsoft Research Paraphrase Corpus (MRPC) that consists of 5,801 sentence pairs collected from newswire articles. Each pair is labelled if it is a paraphrase or not by human annotators. [27] [28]

---

[9] "SemEval2014": http://alt.qcri.org/semeval2014/

[10] "GloVe," Stanford: https://nlp.stanford.edu/projects/glove/

[11] "SemEval2014": http://alt.qcri.org/semeval2014/

[12] "Pheme," 2016: https://www.pheme.eu/2016/04/12/pheme-rte-dataset/

Wang et al. [29] proposed (GLUE[13]) General Language Understanding Evaluation benchmark, a tool for evaluating and analyzing the performance of models across a diverse range of existing NLU tasks based on NLI. Moreover, Wang et al [30] proposed SuperGlue [31] that is an improvement on Glue by having more challenging tasks, more diverse task formats and so on. Glue and SuperGlue contains The QNLI (Question-answering NLI) dataset that is a Natural Language Inference dataset automatically derived from the Stanford Question Answering Dataset v1.1 (SQuAD). SQuAD v1.1 consists of question-paragraph pairs, where one of the sentences in the paragraph (drawn from Wikipedia) contains the answer to the corresponding question (written by an annotator).

Allen Institute for Artificial Intelligence's research created Abductive Natural Language Inference (alphaNLI) [32] that is a common sense benchmark dataset designed to test an AI system's capability to apply abductive reasoning and common sense to form possible explanations for a given set of observations. Formulated as a binary-classification task, the goal is to pick the most plausible explanatory hypothesis given two observations from narrative contexts.

Yuta et al. [33] proposed a dataset for NLI in document-level to support Contract Review process automatically. They simply the problem by modeling it as multi-label classification over spans instead of trying to predict the start and end tokens and they showed that Span NLI BERT outperforms the existing models.

Wang et al [34] solved many NLU tasks by transforming them into NLI task and systematic evaluation on 18 standard NLP tasks shows that it improves the various existing SOTA few-shot learning methods by 12%, and yields competitive few-shot performance with 500 times larger models, such as GPT-3.

Liu et al. [35] proposed RoBERTa that is a BERT tuned model that achieves state-of-the-art results on GLUE, RACE and SQuAD, without multi-task fine-tuning for GLUE or additional data for SQuAD.

He et al. [36] proposed DeBERTa model architecture (Decoding-enhanced BERT with disentangled attention) that improves the BERT and RoBERTa models using two novel techniques(disentangled attention, enhanced mask decoder). Compared to RoBERTa-Large, a DeBERTa model trained on half of the training data performs consistently better on a wide range of NLP tasks, achieving improvements on MNLI by +0.9% (90.2% vs. 91.1%), on SQuAD v2.0 by +2.3% (88.4% vs. 90.7%) and RACE by +3.6% (83.2% vs. 86.8%).

## 2.2. Related Works on Arabic Language

In Arabic language, only few researches were done in RTE domain. Textual entailment in Arabic language faces various challenges due to the features of Arabic language [37] [38] and [39]. One of these challenges is lexical ambiguity, which is the difficulty to process texts with missing diacritics. Another challenge is its richness in synonyms, where more than one-word surface may have the same meaning. In addition, Arabic still lack the large scale handcrafted computational resources that is very practically used in English such as a large WordNet and so on. On the other hand, the lack of large entailment data set caused the lack of deep learning research experiments (only traditional machine learning methods are proposed). Alabbas [40] developed the system ArbTE, to evaluate the existing text entailment techniques when applied to Arabic language. In a next step, Alabbas suggested

---

in [41] extending the basic version of the Tree Edit Distance TED algorithm, to enhance the matching algorithm to identify TE in Arabic. The author also created a publically available data set for Arabic textual entailment ArbTEDS[14] that consists of 618 text-hypothesis pairs collected from Arabic news websites or from annotated pairs collected by hand.

AlKhawaldeh et al. [42] concluded that the Arabic entailment accuracy can be enhanced by resolving negation for entailment relation and analyzing the polarity of the text-hypothesis pair and determining the polarity of the text-hypothesis pair (Positive, Negative or Neutral). They achieved an accuracy of 69% on ArbTEDS data set.

Almarwani et al. [37] applied SVM and Random Forest classifiers to detect RTE in Arabic using word embeddings to overcome the lack of explicit lexical overlap between sentences pairs T and H. They derived word vector representations for about 556K words. Other features used, were: similarity scores, named entities, number of unique instances in T, number of unique instances in H, number of unique instances that are in T but not in H and vice versa, and number of instances that are in both H and T. All features were calculated at token, lemma, and stem levels. The system achieved an accuracy of 76.2% on ArbTEDS data set.

Boudaa et al. [43] used Support Vector Machine algorithm to detect RTE for Arabic language. The following analysis were used in the pre-processing stage: Named Entities, Temporal Expressions, Number/Countable pairs, Ordinary Words (or sequence of ordinary words). They extracted alignment based features to find an optimal weight matching in a weighted bipartite graph. The system achieved an accuracy of 75.84% on ArbTEDS data set.

Khader et al. [39], applied a lexical analysis technique of Textual Entailment for Arabic language. They added a semantic matching approach to enhance the precision of their system. Their lexical analysis is based on calculating word overlap and bigram extraction and matching. They combined semantic matching with word overlap to increase the accuracy of words matching. They achieved a precision of 68%, 58% for both Entails and Not-Entails respectively with an overall recall of 61% on ArbTEDS data set.

### 3. OUR METHODOLOGY

In this work, we created a data set and propose a system to detect the NLI in Arabic sentences, where target labels are Entailment, Contradiction and Neutral (no semantic relation). Our system consists of three main parts: Text Pre-processing (cleaning, tokenization, stemming), Feature Extraction (Contradiction feature vector and language model vectors) and Machine Learning Model. Figure 1 shows Our Experiments Schema. We will discuss each step in details in the following subsections.

---

[14] Arabic Textual Entailment Dataset: http://www.cs.man.ac.uk/~ramsay/ArabicTE/

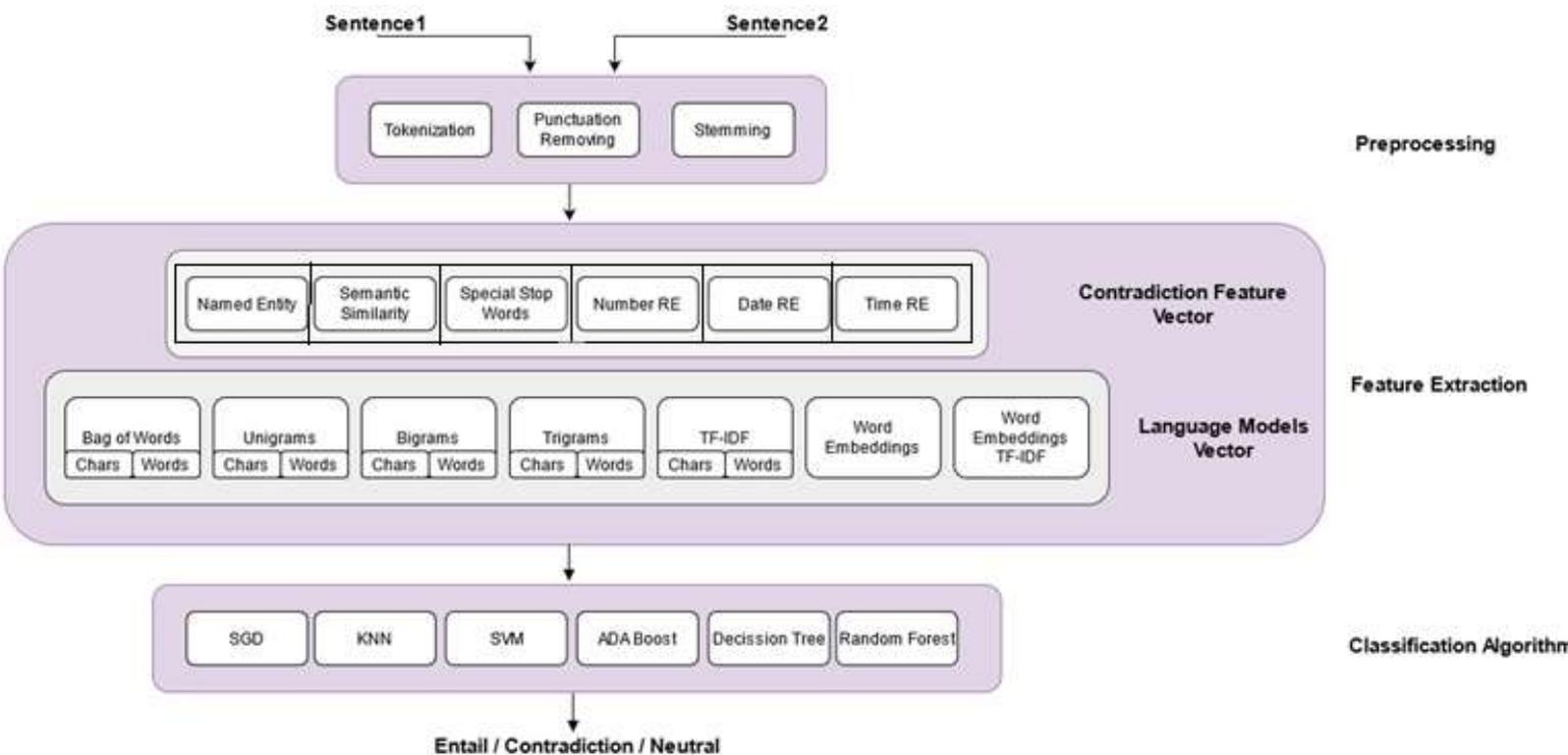

Figure 1 Our Experiments Schema

### 3.1. OUR DATA SET

To the best of our knowledge there is no available Arabic Three-way Natural Language Inference (NLI) data set. In order to build our data set, we started by translating two English RTE data sets: SICK data set [44] which was used in SemEval_2014_Task1[15], and PHEME data set. The SICK data set consists of 10,000 English sentence pairs, each annotated for relatedness in meaning and for entailment relation. PHEME data set, on the other hand, contains 5400 RTE annotated pairs from social media. We named the Arabic automatically translated data sets Ar_SICK and Ar_PHEME respectively.

After automatically translating the two data sets, we selected a subset of the annotated pairs and manually corrected their translations. We augmented this subset with a manually translated/annotated pairs from pre-existed sources. Our final Arabic Natural Language Inference (NLI) data set[16] (ArNLI) contains 6366 pairs divided as (1932 entailment, 1073 contradiction, and 3361 neutral). The data set is collected as following:

- 5948 pairs of AR_SICK data set sentences that were semi-automatically translated and corrected (1714 entailment, 895 contradiction, 3339 neutral pairs)
- 312 pairs of ArbTEDS corpus[17], that we had to re-annotate its sentences from (Entails, Not Entails) classes into our 3-way RTE classes considered in this study (194 entailment, 113 contradiction, 5 neutral pairs).
- 35 pairs of Stanford Real Life contradiction corpus [10], which was manually translated (0 entailment, 35 contradictions, 0 neutral pairs)

---

[15] "semeval2014_task1," 2014. [Online]. Available: https://alt.qcri.org/semeval2014/task1/.

[17]Arabic Textual Entailment Dataset http://www.cs.man.ac.uk/~ramsay/ArabicTE/

- 71 pairs of manually annotated sentences (collected from online websites teaching Arabic contradiction, poems, idioms, paraphrased pairs of Ar_PHEME data set.) with (24 entailments, 30 contradictions, 17 neutral pairs).

The key statistics of our created data set (ArNLI) are shown in Table 1.

| **Data Size** | |
|---|---|
| Training pairs | 5092 |
| Testing pairs | 1274 |
| **Avg. Sentence Length in tokens** | |
| Hypothesis | 6.623 |
| Premise | 7.246 |
| **Max. Sentence Length in tokens** | |
| Hypothesis | 26 |
| Premise | 57 |

Table 1 Key Statistics of ArNLI Dataset

## 3.2. Text Pre-processing

In this step, we first tokenized the sentences and removed all punctuations. To extract the morphological units, we used Snowball Stemmer which is also known as the Porter2 stemming algorithm.

Table 2 presents examples of each step in pre-processing stage

| Stage | Sentence1 | Sentence2 |
|---|---|---|
| | عملنا في هذا البحث على فهم علاقات الاستدلال و استخراجها بين الجمل في جميع اللغات، وليس فقط اللغة العربية.. | عملنا في هذا البحث على اكتشاف علاقات الاستدلال و التناقضات بين الجمل في اللغة العربية فقط لم نعمل على اكتشافها في باقي اللغات ! |
| Tokenization | ['عملنا', 'في', 'هذا', 'البحث', 'على', 'فهم', 'علاقات', 'الاستدلال', 'و', 'استخراجها', 'بين', 'الجمل', 'في', 'جميع', 'اللغات', 'وليس', 'فقط', 'اللغة', 'العربية', '..'] | ['عملنا', 'في', 'هذا', 'البحث', 'على', 'اكتشاف', 'علاقات', 'الاستدلال', 'و', 'التناقضات', 'بين', 'الجمل', 'في', 'اللغة', 'العربية', 'فقط', 'لم', 'نعمل', 'على', 'اكتشافها', 'في', 'باقي', 'اللغات', '!'] |
| Punctuation Removal | ['عملنا', 'في', 'هذا', 'البحث', 'على', 'فهم', 'علاقات', 'الاستدلال', 'و', 'استخراجها', 'بين', 'الجمل', 'في', 'جميع', 'اللغات', 'وليس', 'فقط', 'اللغة', 'العربية'] | ['عملنا', 'في', 'هذا', 'البحث', 'على', 'اكتشاف', 'علاقات', 'الاستدلال', 'و', 'التناقضات', 'بين', 'الجمل', 'في', 'اللغة', 'العربية', 'فقط', 'لم', 'نعمل', 'على', 'اكتشافها', 'في', 'باقي', 'اللغات'] |
| Snowball Stemmer | ['عمل', 'في', 'هذا', 'بحث', 'على', 'فهم', 'علاق', 'استدلال', 'و', 'استخراج', 'بين', 'جمل', 'في', 'جميع', 'اللغ', 'ليس', 'فقط', 'اللغ', 'عرب'] | ['فقط', 'لم', 'نعمل', 'على', 'اكتشاف', 'في', 'باق', 'اللغ', 'عرب', 'استدلال', 'و', 'تناقض', 'بين', 'جمل', 'في', 'لغ', 'عرب', 'على', 'اكتشاف', 'و', 'علاق'] |

Table 2: Example of Output of Each Step In Pre-processing Stage

## 3.3. Feature Extraction

In our proposed model, we used different types of features: Named entity features, WordNet similarity features, Special stopwords feature, Number, Date and Time features. We used different language models, such as TFIDF, N-Grams, and Word Embeddings

### 3.3.1. Contradiction Vector Proposed Features

### A. Arabic Named Entity Features

Two sentences with different named entities cause a contradiction in meaning even they may have almost the same words. For Example, the capital of a country is a specific city that cannot be replaced with another city:

(باريس عاصمة فرنسا) **Contradicts** (ليون عاصمة فرنسا)

(Paris is the capital of France) **Contradicts** (Lyon in the capital of France)

We used AQMAR [27], to detect the named entities in sentences. We encode values to consider three different cases:

- If the same ANEs are used in both sentences.
- If different ANEs are used in both sentences.
- If neither of sentences contains ANE

## B. Semantic Similarity Features

To be able to focus on the concepts (not only on the exact words), we added some semantic features based on WordNet Similarity project. The word can have different meanings according to its context and this has a direct effect on the relations between phrases. Semantic Similarity features are calculated for all words of sentence1 with all words of sentence2. Similarity features are Synonym Words Count, Neutral Words Count, Antonym Words Count. Table 3 presents examples of different relations between sentences in Arabic.

| Relation | Sentence1 | Sentence2 |
|---|---|---|
| No Relation with | شرب الغزال من العين <br> The deer drank from the spring | أعاني من حساسيّة العين <br> I have eye allergy |
| Entailment | اشترى أحمد بيت أمجد <br> Ahmad bought Amjad's house | باع أمجد بيته لأحمد <br> Amjad sold his house to Ahmed |
| Entailment | أفل القمر <br> The Moon sets | أشرقت الشّمس <br> The Sun rises |
| Contradiction | أفل القمر <br> The Moon sets | غربت الشّمس <br> The Sun sets |
| Contradiction | لم ينطق بكلمة <br> He did not say a word. | قال ماما لقد عدت للمنزل <br> He said, Mum I am home. |

Table 3: Different Relation Examples Between Sentences

## C. Arabic Special Stopwords Features

Some Arabic stopwords affects the meaning of sentence, and thus must be considered when studying entailments. In contradiction, for example, negations such as (ليس, لا, ما), and exceptions such as (عدا, سوى, إلا) can alter the results. Moreover, some negation words would mean confirmation if they come together with a negation word, such as (لا, إلا). Table4 presents some examples of contradiction and entailment using stopwords.

| Relation | Sentence1 | Sentence2 |
|---|---|---|
| Contradiction | لا إله <br> No God | لا إله إلا الله <br> No God except Allah |
| Entailment | لا يعلم المستقبل إلا الله <br> No one know the future except Allah | الله يعلم المستقبل <br> Allah knows the future |

Table 4 Relation examples using stopwords

In our system, each special stopword will be extracted, then we encode feature values to consider three different cases:

- if a special stopword exists in one sentence
- if a special stopword exists in both sentence
- if a special stopword does not exist in neither of sentences.

### D. Number, Date and Time features

We extract features concerning Number, Date and Time using Regular Expressions to detect patterns. We also take into consideration Arabic words that compares quantities (ex: حوالي/nearly, ينقص/less than, يزيد عن/more than, etc.). Table 5 shows some examples of contradiction and entailment relations based on these features.

| Relation | Sentence1 | Sentence2 |
|---|---|---|
| Entailment | بلغ عدد ضحايا زلزال اليابان 60 قتيلاً<br>The number of Japan earthquake victims reached 60 | زلزال في اليابان وما يزيد عن 50 قتيلاً<br>Earthquake in Japan and more than 50 killed |
| Entailment | مقتل 3 أطفال و 5 نساء في زلزال اليابان<br>3 children and 5 women killed in Japan earthquake | مقتل 8 أشخاص في زلزال اليابان<br>8 people killed in Japan earthquake |
| Contradiction | ولد خالد عام 1987<br>Khaled was born in 1987 | ولد خالد عام 1990<br>Khaled was born in 1990 |
| Contradiction | بلغ عدد ضحايا زلزال في اليابان 60 قتيلاً<br>The number of Japan earthquake victims reached 60 | زلزال في اليابان وما يقل عن 50 قتيل<br>Earthquake in Japan and less than 50 killed |

*Table 5* Relation *E*xamples using Number, Date and Time

In our system, we create a vector to encode each type of these regular expressions types (number, time, date) into two values to consider two different cases:

- if the quantity value of regular expression is NOT the same in both sentences.
- if the same quantity value of regular expression is in both sentences.

### 3.3.2. Language Models

In this work, we have used different language models to represent the pairs of sentences. We compared the results of the following language models:

- **Bag of Words**: A bag-of-words means an unordered set of words, ignoring their exact position. The simplest bag-of-words approach represents the context of a target word by a vector of features, each binary feature indicating whether a vocabulary word w does or doesn't occur in the context. Bag-of-word features are effective at capturing the general topic of the discourse in which the target word has occurred. This, in turn, tends to identify senses of a word that are specific to certain domains [45] In this work, we extracted bag of words based on *words* vs. *chars* in each sentence of pairs.

- **N-grams**: An n-gram is a continuous sequence of n items from the given sequence of text or speech data. N-grams models assign a conditional probability to possible next words or assign a joint probability to an entire sentence. N-grams are essential in any task in which we have to identify words in noisy, ambiguous input. [45] In this work, we extracted unigrams, Bi-grams, Tri-grams for *words* vs. *chars* in each sentence of pairs.

- **TF-IDF** (term frequency–inverse document frequency) is a term weighting scheme commonly used to represent textual documents as vectors (for purposes of classification, clustering, visualization, retrieval, etc.). Let T = {t1,…, tn} be the set of all terms occurring in the document corpus under consideration. Then a document di is represented by a n-dimensional real-valued vector xi = (xi1,…, xin) with one component for each possible term from T. The most common TF–IDF weighting is defined by $x_{ij} = TF_i \cdot IDF_j \cdot (\sum_j (TF_{ij} IDF_j)^2)^{-1/2}$ [46] In this work, we extracted TF-IDF based on *words* vs. *chars* in each sentence of pairs.

- **Word Embeddings** low-dimensional word vector that encode semantic meanings about the words [47] In this work, we created a word2vec models using Genism implementation. The training was done using 50% of translated sentences from SICK, PHEME data sets.

## 3.4. Classification Models

In order to detect the relation type (Contradiction, Entailment, or Neutral) between two sentences, we used different machine learning classifiers and compared their results. The algorithms used were Support Vector Machine (SVM) [48], Stochastic gradient descent (SGD) [49], Decision Tree(DT) [50], ADA Boost Classifier [51], K-Nearest Neighbors (KNN) [52], Random Forest [53].

## 4. EVALUATION & RESULTS

We evaluated our proposed solution on our created data set (ArNLI) and on both Ar_SICK and Ar_PHEME data sets. Each data set was divided into training and testing sets as 80% and 20% respectively. Table 6 presents the results of applying the different algorithms on ArNLI data set.

| | | | SVM | SGD | DT | ADA | KNN | RF |
|---|---|---|---|---|---|---|---|---|
| **TFIDF** | | Char | 0.65 | 0.65 | 0.59 | 0.52 | 0.52 | 0.73 |
| | | Word | 0.63 | 0.65 | 0.57 | 0.51 | 0.57 | 0.7 |
| | | Union | 0.65 | 0.63 | 0.59 | 0.52 | 0.56 | 0.73 |
| **Bag of Words** | | Chars | 0.64 | 0.57 | 0.59 | 0.53 | 0.56 | **0.75** |
| | | Words | 0.61 | 0.65 | 0.57 | 0.55 | 0.6 | 0.71 |
| **N-Grams** | Words | Unigram | 0.62 | 0.61 | 0.57 | 0.54 | 0.51 | 0.71 |
| | | Bigram | 0.59 | 0.62 | 0.59 | 0.54 | 0.51 | 0.72 |
| | | Trigram | 0.58 | 0.52 | 0.59 | 0.55 | 0.57 | **0.75** |
| | Chars | Unigram | 0.62 | 0.63 | 0.57 | 0.54 | 0.52 | 0.72 |
| | | Bigram | 0.62 | 0.62 | 0.57 | 0.54 | 0.54 | 0.65 |
| | | Trigram | 0.6 | 0.62 | 0.57 | 0.52 | 0.54 | 0.61 |
| **W2Vec** | | word2vec | 0.57 | 0.59 | 0.57 | 0.52 | 0.53 | 0.67 |
| | | word2vec TF-IDF | 0.57 | 0.55 | 0.56 | 0.55 | 0.59 | 0.66 |

Table 6: Results of experiments on ArNLI

Experiments show that Random Forest achieved the best results on ArNLI data sets with an accuracy of 0.75. As for language models used in feature extraction, we found that the best results are achieved by combining Tri-grams of words vector with contradiction vector or combining bag-of-words of chars vector with contradiction vector.

We applied the different experiments on the automatically translated data sets Ar_PHEME and Ar_SICK. Tables 7 and Table 8 show the accuracy results achieved by our experiments on both data sets respectively.

| | | | SVM | SGD | DT | ADA | KNN | RF |
|---|---|---|---|---|---|---|---|---|
| **TFIDF** | | Char | 0.91 | 0.84 | 0.58 | 0.77 | 0.93 | **1** |
| | | Word | 0.89 | 0.85 | 0.52 | 0.78 | 0.92 | **1** |
| | | Union | 0.89 | 0.84 | 0.58 | 0.77 | 0.93 | **1** |
| **Bag of Words** | | Chars | 0.94 | 0.88 | 0.57 | 0.78 | 0.91 | **1** |
| | | Words | 0.91 | 0.87 | 0.52 | 0.78 | 0.93 | **1** |
| **N-Grams** | Words | Unigram | 0.63 | 0.6 | 0.53 | 0.64 | 0.88 | **1** |
| | | Bigram | 0.9 | 0.87 | 0.57 | 0.76 | 0.92 | **1** |
| | | Trigram | 0.92 | 0.88 | 0.52 | 0.76 | 0.91 | **1** |
| | Chars | Unigram | 0.63 | 0.58 | 0.53 | 0.64 | 0.88 | **1** |
| | | Bigram | 0.9 | 0.87 | 0.57 | 0.76 | 0.92 | **1** |
| | | Trigram | 0.92 | 0.89 | 0.52 | 0.76 | 0.91 | **1** |
| **W2Vec** | | word2vec | 0.47 | 0.43 | 0.46 | 0.56 | 0.8 | **1** |
| | | word2vec TF-IDF | 0.49 | 0.46 | 0.5 | 0.58 | 0.8 | 0.99 |

Table 7 Results Accuracy on AR_PHEME Dataset

In Table 6, we notice that the best results (100% accuracy) are achieved using Random Forest on the translated Ar_PHEME data set. This can be justified by the fact that PHEME data set has many repetitions, and that PHEME sentences are initially news headlines which are lexically contradicted (such in "ten people are died in Airbus crash" and "no one is died in the Airbus crash"), and thus can be easily detected.

When applying the different experiments on the automatically translated data set Ar_SICK, we notice that the best results (an accuracy of 66%) has been achieved using ADA algorithm

| | | | SVM | SGD | DT | ADA | KNN | RF |
|---|---|---|---|---|---|---|---|---|
| **TFIDF** | | Char | 0.58 | 0.56 | 0.58 | 0.53 | 0.59 | 0.53 |
| | | Word | 0.52 | 0.48 | 0.64 | 0.55 | 0.58 | 0.52 |
| | | Union | 0.52 | 0.54 | 0.63 | 0.53 | 0.58 | 0.56 |
| **Bag of Words** | | Chars | 0.58 | 0.6 | 0.6 | 0.66 | 0.52 | 0.57 |
| | | Words | 0.54 | 0.57 | 0.6 | 0.52 | 0.52 | 0.48 |
| **N-Grams** | Words | Unigram | 0.58 | 0.57 | 0.6 | 0.56 | 0.52 | 0.54 |
| | | Bigram | 0.52 | 0.57 | 0.64 | 0.55 | 0.54 | 0.57 |
| | | Trigram | 0.52 | 0.58 | 0.63 | 0.63 | 0.58 | 0.57 |
| | Chars | Unigram | 0.58 | 0.6 | 0.55 | 0.65 | 0.52 | 0.58 |
| | | Bigram | 0.57 | 0.57 | 0.56 | **0.66** | 0.58 | 0.6 |
| | | Trigram | 0.5 | 0.58 | 0.55 | 0.63 | 0.58 | 0.6 |
| **W2Vec** | | word2vec | 0.53 | 0.6 | 0.53 | **0.66** | 0.57 | 0.57 |
| | | word2vec TF-IDF | 0.52 | 0.57 | 0.52 | **0.66** | 0.5 | 0.58 |

Table 8 Results Accuracy on AR_SICK Dataset

with both W2Vec or Bi-gram on char level language model (see Table 8).

When comparing the results of the different data sets, we remark that worst results were achieved on Ar_SICK data set (66%). We can justify that by the fact that this data set contains many pairs with semantic abstraction level, and the automatic translation step has changed the semantics of one or both sentences, making the original label not valid.

Table 9 presents few examples of entailment that our system failed to detect in Ar_SICK dataset.

| Sentence1 | Sentence2 | Automatic Translation of Sentence1 | Automatic Translation of Sentence2 | Original Label |
|---|---|---|---|---|
| A dog is rolling on the ground | A dog is sleeping on the ground | كلب هو المتداول على الأرض (بدلاً من كلب يتدحرج على الأرض) | كلب نائم على الأرض | NEUTRAL |
| A horse is being ridden by a person | A person is riding a horse | ويجري تعصف بها حصان من قبل شخص (بدلاً من حصان يركبه شخص) | شخص يركب الخيل | ENTAILMENT |
| A person is tearing sheets | A man is cutting a paper | يكون الشخص ورقة تمزيق (بدلاً من يقوم شخص بتمزيق الملاءات) | رجل يقطع ورقة | NEUTRAL |
| There is no woman cutting broccoli | A woman is cutting broccoli | لا يوجد البروكلي قطع امرأة (بدلاً من لا توجد امرأة تقطع البروكلي) | امرأة تقطع البروكلي | CONTRADICTION |

Table 9 Examples of Translation spoiling semantics

Average results are achieved on our data set (ArNLI) (75%), as it includes different types of contractions, with different levels of semantics.

Figures 1,2,3 show comparisons of results of all algorithms on translation of PHEME Dataset, translation of SemEval2014Task1 Dataset, and ArNLI Dataset respectively.

Figures 4,5,6,7 show comparisons of best results on the three used datasets (translation of PHEME Dataset, translation of SemEval2014Task1 Dataset, and ArNLI Dataset) using Support Vector Machine (SVM), ADA Boost, Stochastic Gradient Descent (SGD), and Random Forest respectively.

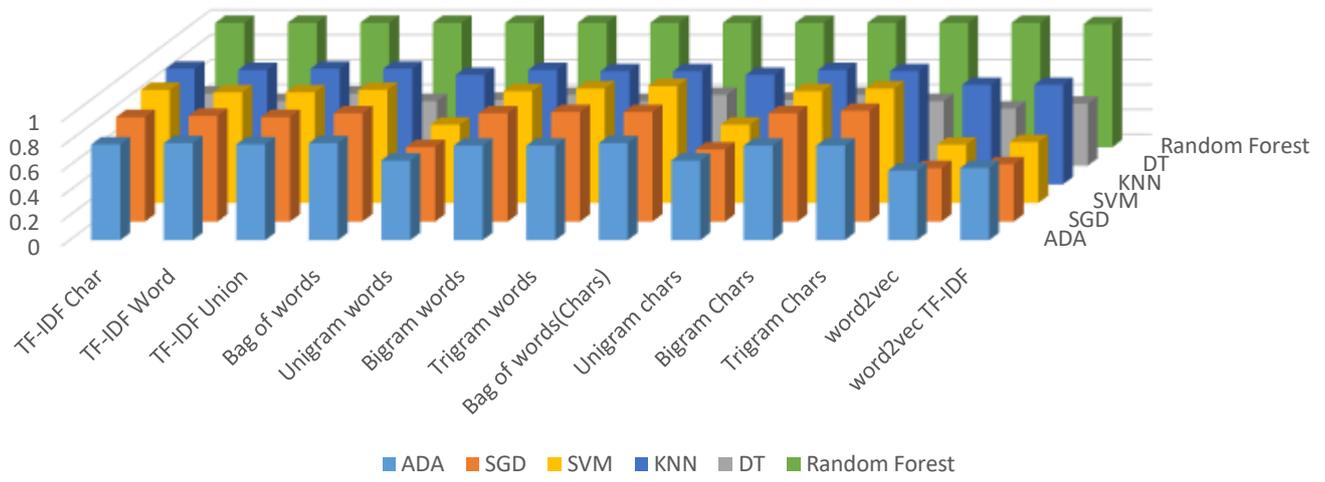

Figure 1 Results on Translation of PHEME Dataset

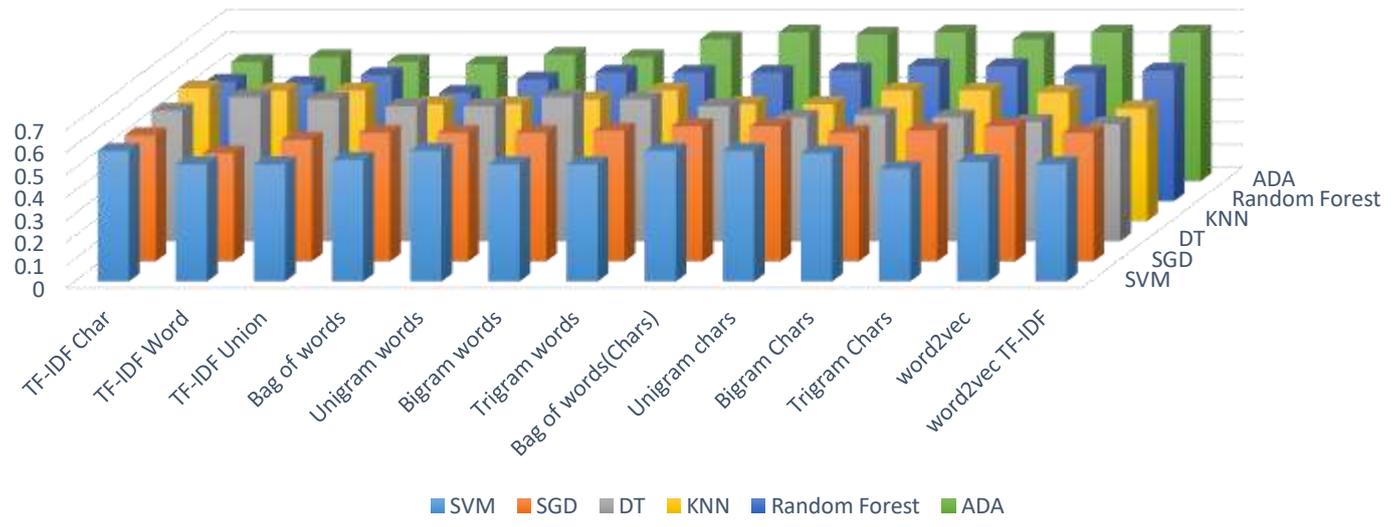

Figure 2 Results on Translation of SemEval2014Task1 Dataset

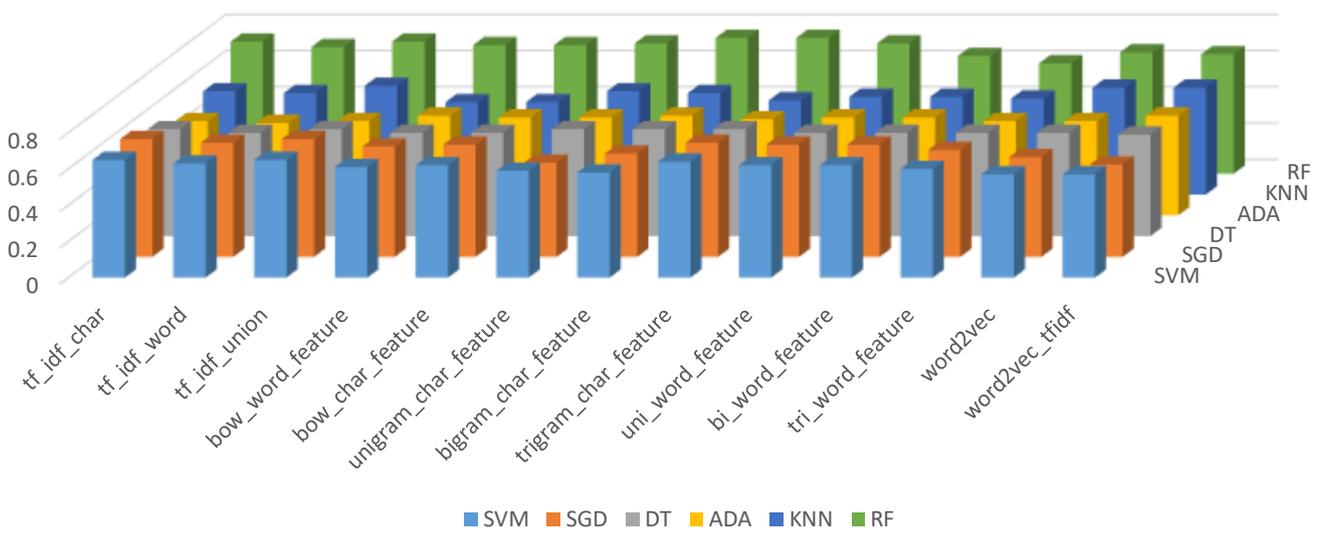

Figure 3 Results on ArNLI Dataset

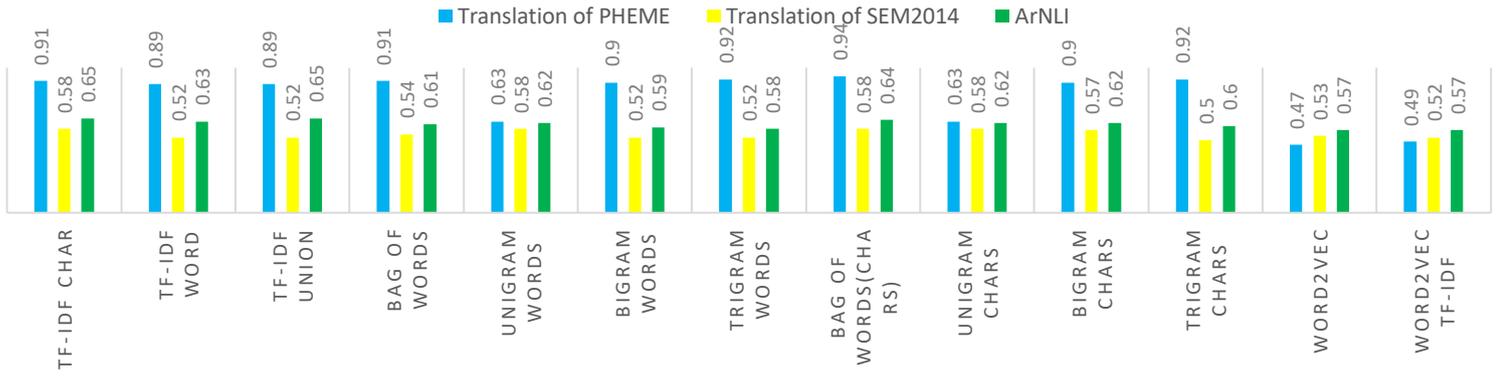

Figure 4 Results of SVM

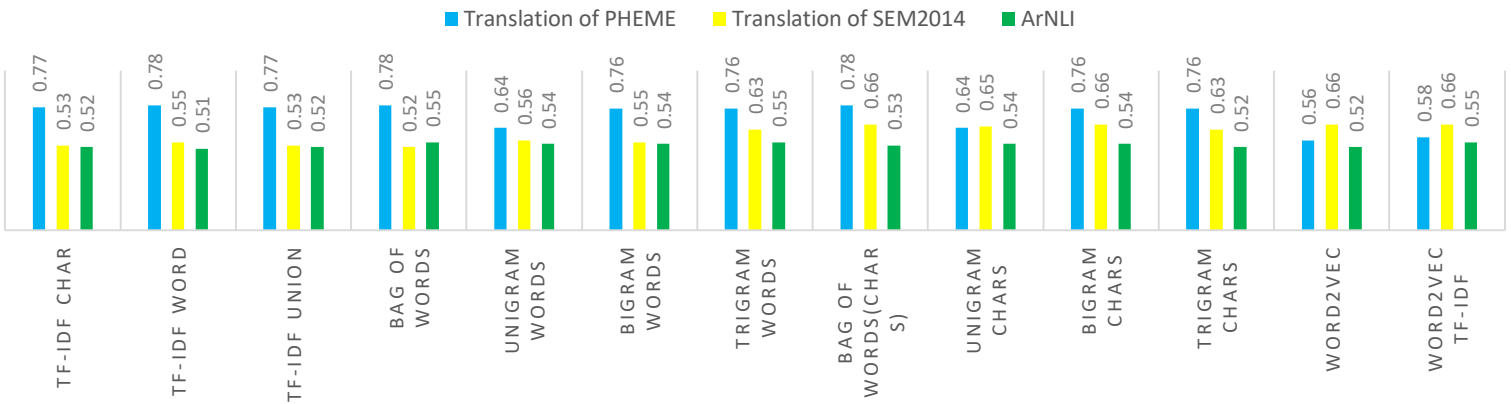

Figure 5 Results of ADA

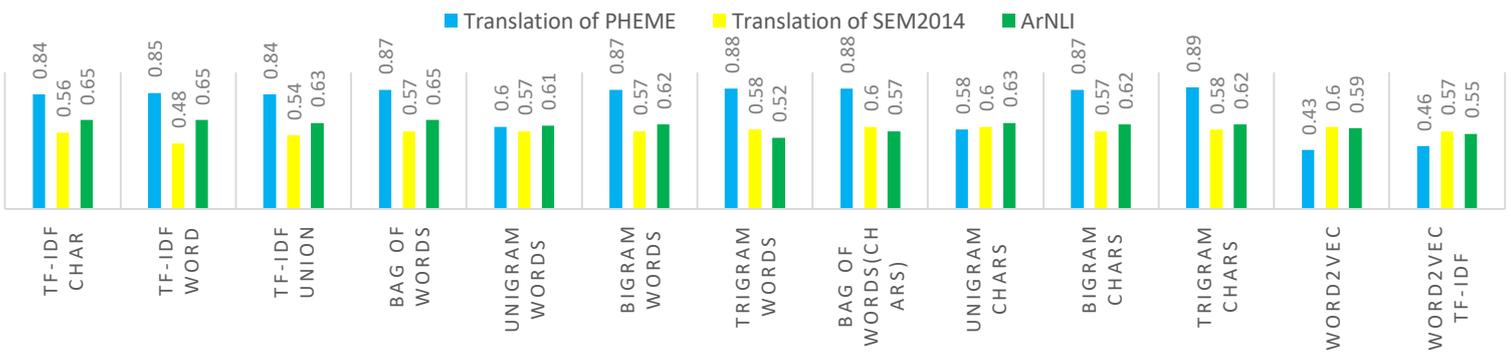

Figure 6 Results of SGD

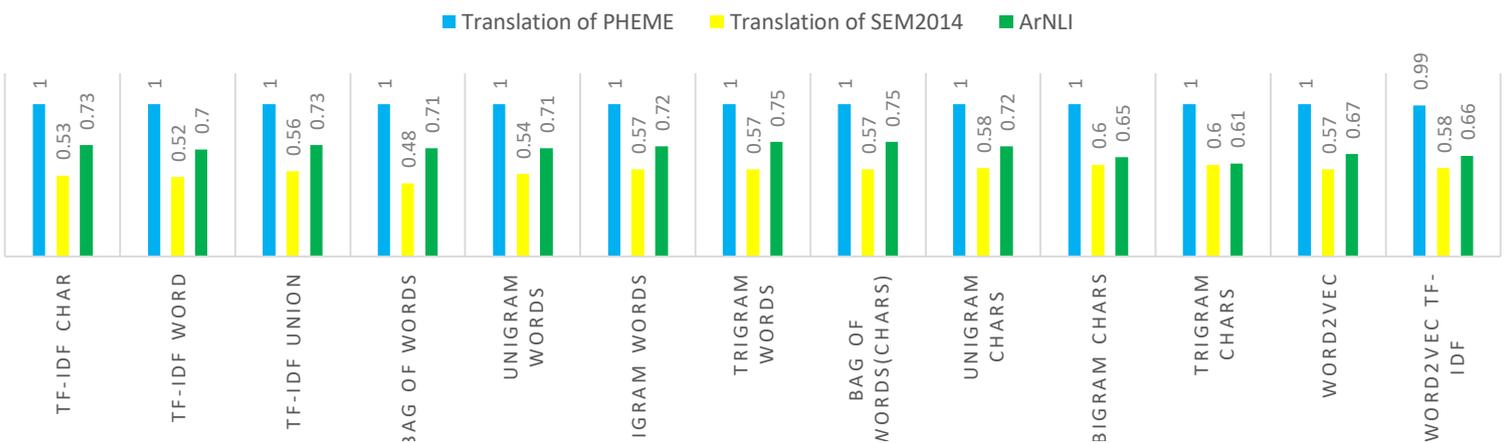

Figure 7 Results of Random Forest

## 5. CONCLUSION

Detecting entailment relations between statements is very essential and challenging NLP task, especially contradiction detection, which can really optimize the core of many NLP applications. Arabic language suffers from low resources in NLI detection, only small data set is available, so no deep learning solutions were proposed in this domain before. In this paper, we presented our semi-automatically created data set ArNLI that contains more than 12k sentences. We automatically translated the English PHEME and SICK data sets. We have made some basic experiments to detect entailments in Arabic language, inspired by Stanford proposed solutions on English language. We applied these experiments on our created data set ArNLI, and compared the results with the translated PHEME and SICK, as the lack of benchmarks in Arabic language. Best results of accuracy of 0.75 on ArNLI dataset were achieved using Random Forest classifier and feature vector containing combination of Tri-grams of words vector with contradiction vector or combination bag-of-words of chars vector with contradiction vector.

In a future step, we intend to augment our data set and perform different experiments using different embeddings, different transformers and different deep learning algorithms. Moreover, we would like to apply NLI as part of other important NLP tasks such as sarcasm detection and machine reading.

### ABBREVIATIONS

**NLI:** Natural Language Inference

**RTE:** Recognize Textual Entailment

## DECLARATIONS

### ETHICS APPROVAL AND CONSENT TO PARTICIPATE

The authors Ethics approval and consent to participate.

### CONSENT FOR PUBLICATION

The authors consent for publication.

### AVAILABILITY OF DATA AND MATERIALS

The data set created in this research is available in the following repository
https://github.com/Khloud-AL/ArNLI

### COMPETING INTERESTS

The authors declare that they have no competing interests.

**FUNDING**

The authors declare that they have no funding.